\pdfoutput=1

\documentclass[11pt]{article}

\usepackage{EMNLP2023}

\usepackage{multirow}
\usepackage{times}
\usepackage{latexsym}

\usepackage[T1]{fontenc}

\usepackage[utf8]{inputenc}

\usepackage{microtype}
\usepackage{graphicx}
\usepackage{inconsolata}
\usepackage{booktabs}
\usepackage{tablefootnote}
\usepackage{lipsum} 

\title{Mini Minds: Exploring Bebeshka and Zlata Baby Models
}

\author{
    ~\textbf{Irina Proskurina}, 
    ~\textbf{Guillaume Metzler},
    ~\textbf{Julien Velcin} \\
    Universit{\'e} de Lyon, Lyon 2, ERIC UR 3083, France\\ 
    \textbf{Correspondence:} \href{mailto:Irina.Proskurina@univ-lyon2.fr}{Irina.Proskurina@univ-lyon2.fr}
}

\begin{document}
\maketitle
\begin{abstract}
In this paper, we describe the University of Lyon 2 submission to the \textsc{Strict-Small} track of the BabyLM competition.
The shared task is created with an emphasis on small-scale language modelling from scratch on limited-size data and human language acquisition. 
Dataset released for \textsc{Strict-Small} track has 10M words, which is comparable to children's vocabulary size. 
We approach the task with an architecture search, minimizing masked language modelling loss on the data of the shared task.
Having found an optimal configuration, we introduce two small-size language models (LMs) that were submitted for evaluation, a 4-layer encoder with 8 attention heads and a 6-layer decoder model with 12 heads which we term Bebeshka and Zlata, respectively.
Despite being half the scale of the baseline LMs, our proposed models achieve comparable performance.
We further explore the applicability of small-scale language models in tasks involving moral judgment, aligning their predictions with human values.
These findings highlight the potential of compact LMs in addressing practical language understanding tasks. 
We make our code and models publicly available.\footnote{\url{https://github.com/upunaprosk/small-language-models}}

\end{abstract}

\section{Introduction}
LMs accurately encode language-specific phenomena required for natural language understanding and generating coherent continuation of text. 
LMs gain language understanding about morphosyntax and grammar from large corpora during pre-training. However, they demonstrate partial functional linguistic competence when applying grammatical knowledge to novel expressions at inference time, which is caused by memorising the most occurring linguistic patterns from the training corpus and limited generalization ability of learnt linguistic representations~\cite{wu-etal-2022-causal,tucker-etal-2022-syntax, mahowald2023dissociating}.

Recent pre-training dynamics studies revealed that the performance of LMs can be seen as a function of training corpus vocabulary: (1) grammatical knowledge improves with the expansion of the pre-training data vocabulary~\cite{van-schijndel-etal-2019-quantity} and (2) small-scale LMs can perform on par with RoBERTa if the vocabulary of used tokenizer is close to the actual human and even child's vocabulary~\cite{liu2019roberta}. 

In this paper, we introduce small-scale LMs with an architecture optimized for the \textsc{Strict-Small} track data of BabyLM competition~\cite{warstadt-et-al-2023-babylm}.
Our objective is to estimate the general performance and capabilities of shallow LMs in downstream tasks beyond the ones suggested in the evaluation pipeline of shared task.
That was achieved through two main contributions.\\
\textit{Contribution 1}. 
We determine an optimal architecture of encoder-based LMs using the Tree-structured Parzen Estimator algorithm and minimal perplexity as a minimizing objective function. 
Our parameter search results suggest that optimal LMs have a ratio of attention heads to layers around 2, while the ratio of previously tested and existing LMs at their base configuration is equal to one.\\
We introduce new small-scale LMs submitted to the shared task: (\emph{i}) 4-layer encoder Bebeshka\footnote{A word used to call a baby in a range of South and East Slavic languages.} and (\emph{ii}) 6-layer decoder Zlata.\footnote{From Zlato (``Golden sweetheart'') used to call babies in West and East Slavic languages.} 
The parameters of the models are presented in \autoref{tab:model_architecture}.
Our LMs perform on par with the shared task baselines, while they are half the size of those.\\
\begin{table*}[h!]
\centering 
\footnotesize
\begin{tabular}{lcccc} 
\toprule
\textbf{Parameter} & \textbf{RoBERTa} & \textbf{Bebeshka}  &  \textbf{GPT-2} & \textbf{Zlata}\\
\midrule
Pre-training objective & MLM  & MLM & CLM & CLM\\
Vocabulary size & 50K & 8K & 50K & 30K\\ 
\midrule
\#Parameters & 125M &  16M &  345M & 66M \\
Positional embedding type &  absolute & rel. key query &  absolute  & absolute \\
Maximum sequence length  & 512 & 128  & 1024 & 1024 \\
($L$, $A$, $H$, $F$)  &  (12, 12, 64, 3072) & (4, 8, 70, 1412) & (24, 16, 64, 4x 1024) & (6, 12, 64, 4x 768) \\
Activation function  & GELU & New GELU  & New GELU & GELU \\
Dropout probability  &  0.1 & 0.15 & 0.1 & 0.2\\
Attention dropout  &  0.1 & 0.3 & 0.1 & 0.2\\
\midrule
Processing  &  1024x V100 & 4x IPU-M2000& 64x V100 & 4x IPU-M2000\\
Processing time & 1 day & 4h & >30 days & 6h\\
Epochs  &  >40  & 10 & >40 & 10\\
\bottomrule
\end{tabular}
\caption{
Model configurations and pre-training details of Bebeshka and Zlata LMs compared to RoBERTa-base and GPT-2 medium.
Our LMs have configurations of optimal architecture determined with an architecture search~(\S\ref{sec:model_selection}). 
GPT-2 official training information has not been publicly disclosed; we report GPT-2 pre-training hardware details when using model parallelism specified by \citealp{shoeybi2019megatron}.
We use Graphcore Intelligence Processing Units (IPUs) for pre-training our LMs (\citealp{jia2019dissecting} provide a detailed review on IPUs).
MLM=Masked Language modelling, CLM=Causal Language modelling,
$L$=Layers, $A$=Attention heads, $H$=Hidden size per head, $F$=Feed-forward (intermediary) layer size.
}
\label{tab:model_architecture}
\end{table*}
~\textit{Contribution 2}.
We investigate the alignment of small-scale LMs predictions with shared human values in the context of moral judgment tasks.
We find that shallow LMs, yet trained on limited corpora, perform on par with base LMs in commonsense morality scenarios, and, surprisingly outperforming existing baselines in such tasks as virtue and justice assessment.  To the best of our knowledge, our work represents one of the earliest attempts to investigate how predictions made by tiny language models trained on a developmentally plausible corpus correlate with human-shared values. 

This paper has the following structure. After a short section dedicated to related work (\S\ref{sec:related}), we first describe tokenizer training~(\S\ref{sec:vocabulary}), architecture search results and optimal model selection~(\S\ref{sec:model_selection}), and the final architecture of the pre-trained LMs~(\S\ref{sec:model_pretraining}). 
Then, we present scores on datasets included in the shared task (\S\ref{sec:exp_results}), and we present ethics evaluation results (\S\ref{sec:moral_judjement}).

\section{Related Work}
\label{sec:related}
Recent large LMs found applications in many NLP tasks, such as grammatical correction, text completion, and question answering; yet, their usage is constrained by their computational cost. 
Previous works reduce the model size and inference time with knowledge distillation, parameter quantization and other compression techniques \cite{sanh2019distilbert, yao-etal-2021-adapt, tao-etal-2022-compression}.
Other studies investigated the relationship between model parameter count and performance.
\citealp{kaplan2020scaling} has introduced scaling laws, showing the power-law dependency between perplexity and the model size, as well as between the training loss and dataset size.
The paradigm of scaling laws further formed the basis for recent research examining the behaviour of LMs at a small scale \cite{fedus2022switch, fu2023specializing}. 
For instance, \citealp{puvis-de-chavannes-etal-2021-hyperparameter} presented results of \textit{Neural Architecture Search} in limited parameter space, suggesting that optimal LMs are smaller than the existing base configurations.

In parallel, there is numerous research focusing on the efficiency of dataset size, vocabulary and representation that can help to reduce computation cost by minimizing the training steps \cite{van-schijndel-etal-2019-quantity, huebner-etal-2021-babyberta, schick-schutze-2021-just, warstadt2022artificial}. 
\citealp{van-schijndel-etal-2019-quantity} have demonstrated that LMs trained on a small-volume corpus can reach human performance under some grammatical knowledge evaluation scenarios, questioning the necessity of large datasets for pre-training.
\citealp{huebner-etal-2021-babyberta} introduced a small encoder-based LM BabyBERTa with 5M parameters and showcased the efficiency of small training data; that work bridged the gap between earlier studies on model size reduction and optimal data size.

The aforementioned related works mainly analyse the difference between compact LMs and their larger counterparts with throughput time measures and performance on GLUE benchmark \cite{wang-etal-2018-glue}. In this paper, we evaluate LMs at a small scale trained on a 10M size dataset of BabyLM shared tasks and try to complement existing research with additional evaluation on moral judgment tasks. 
The decision to focus on the moral judgment task is driven by recent studies that reveal human-like biases in the moral acceptability judgments made by large language models trained on extensive corpora~\cite{schramowski2022large}. 
This paper complements existing research by conducting a moral judgment evaluation for small language models.

\section{Methodology}
\label{sec:methodology}
\begin{table*}[th!]
\centering 
\scriptsize
\begin{tabular}{lccc} 
\toprule
\textbf{Parameter} & \textbf{Search range}  &  \textbf{10\% Best runs Mean} & \textbf{10\% Worst runs Mean}\\
\midrule
Positional embedding type & (rel. key,
rel. key query, absolute) &  rel. key query & absolute \\
\# Hidden layers & [1-12] & 6.2 & 10.9\\ 
\# Attention heads &  [1-18] & 11.9 &  7.1\\
Hidden size per head  & [1-100] & 81.6 & 64.1\\
Feed-forward layer size & [1-3072] & 1446.3 & 2034.5\\
Activation function  &  (New GELU, GELU, SiLU, ReLU)& New GELU & ReLU \\
Dropout probability  &  [0.1-1.0] & 0.16 & 0.63\\
Attention dropout  &  [0.1-1.0]& 0.33 & 0.70\\
\midrule
Avg. perplexity  & - & 24.53 & 992.27\\
\bottomrule
\end{tabular}
\caption{Parameter search space of Optuna study for pre-training encoder LMs on \textsc{Strict-Small} corpus and mean parameter values across 10 best and worst runs sorted by the perplexity. For non-numerical parameters, we report the most common parameter values among study runs.}
\label{tab:model_selection}
\end{table*}
We follow pre-training tasks of RoBERTa \cite{liu2019roberta} and GPT-2 \cite{radford2019language} and refer to these as the architecture baselines in this section. 
We train Bebeshka\footnote{\url{https://huggingface.co/iproskurina/bebeshka}} and Zlata\footnote{\url{https://huggingface.co/iproskurina/zlata}} with masked language and causal language modelling objectives, respectively, and compare their vocabularies and architectures with the baselines.
\subsection{Vocabulary}
\label{sec:vocabulary}

\paragraph{Training Data}
We use data provided within the \textsc{Strict-Small} track of the shared task. 
We report statistics of the training corpus in \autoref{tab:corpora-summary} (\autoref{sec:experimental_frame}). 
The transcribed speech, extracted from recordings of casual speech addressed to children and educational movie subtitles, makes up the bulk of the corpus. 
The average length of the texts is around 30 tokens; considering that and the maximum text length, we lower the maximum sequence length from the base 512 to 128 tokens for the configuration of our LMs. 
\paragraph{Input Representation}

We follow tokenization models of the baselines (GPT-2, RoBERTa) and BabyBERTa~\cite{huebner-etal-2021-babyberta} and use byte-level Byte-Pair Encoding (BPE) algorithm ~\cite{sennrich-etal-2016-neural}; that is, a tokenization method based on iterative merging of the most occurring bytes pairs in a further shared vocabulary.
For the encoder Bebeshka, we build a case-insensitive vocabulary\footnote{We use BPE implementation available under HuggingFace Tokenizers library~\cite{Moi_HuggingFace_s_Tokenizers_2023}.} of size 8K. 
We find a few mismatches between Bebeshka and RoBERTa tokenization and provide more details in \autoref{sec:tokenization_tests}.
The decoder Zlata has a 30K vocabulary constructed with default parameter settings of Tokenizers trainer;\footnote{\url{https://github.com/huggingface/tokenizers}} that value also allows for bypassing the inclusion of
onomatopoeic words that prevail in some transcribed texts of the shared task data. 
\subsection{Model Selection}
\label{sec:model_selection}
To determine an optimal configuration of encoder LM, we use an Optuna-implemented Bayesian optimization algorithm \cite{10.1145/3292500.3330701} and tune parameters listed in \autoref{tab:model_selection} that determine the architecture.
The upper bounds of the numerical parameters in a search space are chosen in accordance with the base RoBERTa configuration.
We set the lower bounds to 1, ensuring a thorough exploration of architectural variations to find the optimal configuration for the masked language modelling task.
Optuna features efficient implementation of optimization algorithms; in our optimization study, we use a standard Tree-structured Parzen Estimator (TPE) algorithm, which uses tree-structured representations and Parzen windows for modelling the probability distributions of hyper-parameters and their density estimation. 
We use TPE to sample parameter values from the search space and an automated early-stopping based on pruning runs with an intermediary perplexity higher than the median of preceding runs.

We set masked language modelling loss (perplexity) of RoBERTa  initialized with the TPE sampled configuration parameters as a minimizing objective function. 
The perplexity is calculated on the \textsc{Strict-Small} validation set after training the model for 10 epochs on written English texts sample (Gutenberg and Children's Book Test corpora and Wikipedia) from the training BabyLM corpus (see \autoref{tab:corpora-summary}). 
We choose a corpus sample to reduce parameter search executing time since dataset size directly impacts an LM training time at each optimization step.
We manually found that training on written texts yields a better score.
Optimization study with an upper bound of 100 trial runs ran for roughly two days on a single A100 GPU.

\autoref{tab:model_selection} reports parameter search results for the best and worst runs according to perplexity on the validation dataset.

The \textbf{optimal configuration} for encoder LMs can be summarized as follows: (1) the ratio of the number of attention heads to the number of layers fluctuates within the 1.5-2 range, (2) employing relative key query type positional embeddings, (3) the dropping ratio 0.3 for attention probabilities.
We further use these three key configuration attributes to initialize Bebeshka.
Parameters other than positional embeddings type, dropout ratio and the number of layers/heads vary significantly across the top 10\% runs. 
Precisely, all types of activation functions, except for ReLU, appear evenly in the best range.
When it comes to the hidden size per head, it takes values from 65 to 85, with a mean of 81.6.
We also observe a notable deviation of intermediary size from the mean value.
Altogether our results show that the best-performing encoder LMs are smaller than the base configuration of RoBERTa, which aligns with \citealp{puvis-de-chavannes-etal-2021-hyperparameter}.
\begin{table}[t]
\centering
\small
\begin{tabular}{lccccc}
\toprule
\multirow{2}{*}{\textbf{Model}} & \multicolumn{2}{c}{\textbf{Loss}} & \multicolumn{2}{c}{\textbf{Run time}} \\
&  \textbf{Val} & \textbf{Test} &  \textbf{Val} & \textbf{Test} \\
\midrule
\multicolumn{5}{c}{\textbf{MLM}}\\
\midrule
RoBERTa (125M) & \underline{3.72} &  \underline{4.42} & 1519 & 1592 \\
Bebeshka (16M) &  \textbf{3.54} & \textbf{4.30} & \textbf{485} & \textbf{649}\\
\midrule
\multicolumn{5}{c}{\textbf{CLM}} \\
\midrule
OPT (125M) & 7.11 & 7.10 & 1493 & 1567\\
Zlata (66M) & 4.64 & 4.69 & \underline{831} & \underline{869} \\
\bottomrule
\end{tabular}
\caption{\label{tab:train_loss} 
Pre-training objective loss on validation and test data of Bebeshka and Zlata compared to baseline models and average run time in seconds.
We run an evaluation of all LMs on the same V100 GPU and use Hugging Face~\href{https://huggingface.co/docs/transformers/main_classes/trainer}{Trainer API} for calculating the scores. 
The best score is in bold, and the second-best score is underlined.
}
\end{table}
\subsection{Model Pre-training}
\label{sec:model_pretraining}

We train our models on 4 Graphcore IPUs with two encoder layers trained on each with mixed precision\footnote{\url{https://www.graphcore.ai/products/ipu}} and use \textsc{Strict-Small} training split.
\autoref{tab:model_architecture} shows the configuration settings of our LMs. 
\paragraph{Bebeshka} The 16M parameters model is based on RoBERTa architecture with determined optimal layer sizes~(\S\ref{sec:model_selection}).
We train Bebeshka on the 10M training corpus of the shared task.
We decrease the probability for selecting masked tokens from standard 15\% to 13.5\%, which is one of the equivalents to set RoBERTa unmasking probability to 0 discussed by \citealp{huebner-etal-2021-babyberta}.

\paragraph{Zlata}
That decoder LM is a light 66M version of GPT-2 with 6 layers trained for 10 epochs on the training \textsc{Strict-Small} data. 
Motivated by the configuration of the best encoder LM, we use the ratio of attention heads to decoder layers equal to 2. 
We explain parameter choice in \autoref{sec:train_details}.

\section{Experiments Results}
\label{sec:exp_results}
In this section, we report the results submitted for the BabyLM shared task. 
LMs discussed in this section are pre-trained on the shared task data, including the baselines. 
We use baselines that were created with existing tokenizers and released by the organizers of the BabyLM competition.\footnote{We also report scores for the version of the model trained with full precision weights, which we dub \href{https://huggingface.co/iproskurina/bebeshka-v2}{Bebeshka-2}. However, we do not discuss those since they were submitted after the leaderboard release.}

\subsection{Pre-training Objective Loss}

We present the evaluation results of our LMs in \autoref{tab:train_loss}, where we compare their performance against the shared task baselines and evaluation runtime. 
While the baselines were trained for 20 epochs, we can observe competitive results by pre-training our small-scale models for ten epochs. 
One of the main advantages of the introduced models lies in their compact size, which makes them more efficient at inference time, even though they do not outperform the baselines by a large margin, which can be seen from the average run time.

\subsection{Linguistic Minimal Pairs}
\begin{figure}
  \centering
  \includegraphics[width=0.95\linewidth]{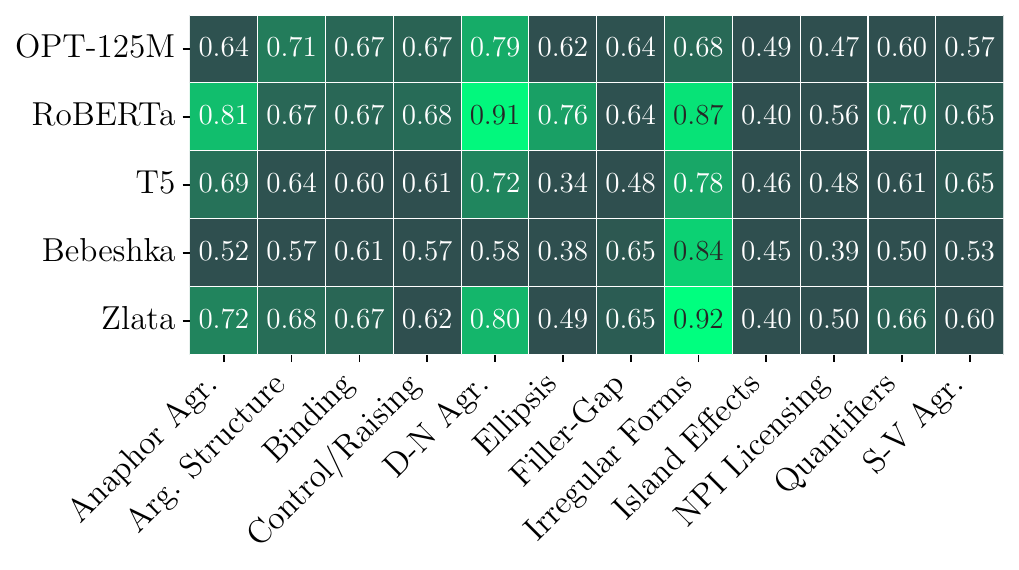}
  \caption{
  Accuracy on BLiMP tasks of our LMs with
RoBERTa-base, OPT-125M, and T5-base
baselines. 
The lighter colours correspond to greater accuracy and, hence, better scores. Morphology: \textit{Anaphor Agr.}, \textit{D-N Agr.}, \textit{Irregular Forms}, \textit{S-V Agr.}. 
Semantics: \textit{NPI Licensing}, \textit{Quantifiers}.
Syntax-Semantics:\textit{Binding}, \textit{Control/Raising}.
The rest phenomena correspond to the Syntax category.}
  \label{fig:blimp_eval}
\end{figure}

\begin{table*}[h!]
\centering
\footnotesize
\begin{tabular}{l|ccccccccccc}
\toprule
\multirow{2}{*}{\textbf{Model}} & \textbf{CoLA}  & \textbf{SST-2}  & \textbf{MRPC} & \textbf{QQP}  & \textbf{MNLI}  & \textbf{MNLI\textsubscript{mm}}  & \textbf{QNLI}  & \textbf{RTE}  & \textbf{BoolQ}  & \textbf{MultiRC}  & \textbf{WSC}  \\
& \textbf{MCC}  & \textbf{Acc.} & \textbf{F1} & \textbf{F1}  & \textbf{Acc.} & \textbf{Acc.}  & \textbf{Acc.} & \textbf{Acc.}  & \textbf{Acc.}  & \textbf{Acc.}  & \textbf{Acc.}  \\
\midrule
OPT & 15.2 & 81.9 & 72.5 & 60.4 & 57.6 & 60.0 & 61.5 & \underline{60.0} & 63.3 & \underline{55.2} & 60.2 \\
RoBERTa & \textbf{25.8} & \textbf{87.0} & \underline{79.2} & \underline{73.7} & \textbf{73.2} & \textbf{74.0} & \textbf{77.0} & \textbf{61.6} & \textbf{66.3} & \textbf{61.4} & \underline{61.4} \\
T5 & 11.3 & 78.1 & \textbf{80.5} & 66.2 & 48.0 & 50.3 & 62.0 & 49.4 & \underline{66.0} & 47.1 & 61.4 \\
\midrule
Bebeshka & 0.11 &  81.3 & 73.5 & 66.4 & 58.7 & 62.0 & 59.0 & 45.4 & 63.9 & 48.7 & 61.4 \\
Zlata & 0.05 & 81.7 & 77.6 & 65.9 & 61.9 & 63.9 & 61.7 & 56.6 & 65.3 & 53.8 & \textbf{61.5} \\
\midrule
Bebeshka-2 & \underline{24.5} &  \underline{83.5} & 77.7 & \textbf{77.3} & \underline{65.4} & \underline{66.9} & \underline{64.0} & 56.6 & 60.2 & 46.9 & 61.4 \\
\bottomrule
\end{tabular}
\caption{
Evaluation results on GLUE and SuperGLUE (BoolQ, MultiRC, WSC) benchmark datasets. 
We report metrics suggested in the shared task evaluation pipeline and baselines.
The best score is in bold, and the second-best score is underlined.
}
\label{tab:glue_results}
\end{table*}
\autoref{fig:blimp_eval} depicts the evaluation results of our LMs on the BLiMP dataset \cite{warstadt-etal-2020-blimp-benchmark} in a zero-shot setting. 
The goal of this evaluation benchmark is to assess a model's ability to distinguish between grammatically acceptable and unacceptable sentences without specific fine-tuning on the task. 
The dataset consists of minimal pairs annotated with a grammatical phenomenon. 
We report detailed LMs accuracy scores across various BLiMP tasks in \autoref{tab:blimp_results}~(\autoref{sec:app_results}).
The general trend is that LMs trained on BabyLM data perform well on minimal pairs with morphological tasks, such as \textit{Irregular Forms} and \textit{Determiner-Noun Agreement}.

Zlata achieves the best accuracy  (92.1\%) on \textit{Irregular Forms} and outperforms OPT-125M baseline on some morphological tasks (\textit{Anaphor Agreement}, \textit{Subject-Verb Agreement}), minimal pairs with a violation in phrasal movements (\textit{Filler Gap}) and  other tasks, such as \textit{NPI Licensing}.
Bebeshka achieves the second-best accuracy (64.7\%) on \textit{Filler Gap} minimal pairs and distinguishes sentences with syntactic errors in pronoun and its antecedent relationship or syntactic islands (\textit{Binding}, \textit{Island Effects}).
The results show that LMs trained on the BabyLM corpus have syntactic and morphology understanding which influences their behaviour on downstream tasks discussed next.

\subsection{GLUE}
\autoref{tab:glue_results} shows results of fine-tuned LMs evaluation on a variety of tasks present in GLUE and SuperGLUE benchmarks.\footnote{Provided datasets within the shared task were filtered according to the vocabulary of BabyLM \textsc{Strict-Small} corpus.}
Submitted to the shared task, Bebeshka and Zlata were fine-tuned for ten epochs on most of the tasks (see \autoref{sec:train_details} for more detail).
The overall trend is that the introduced small-scale encoder Bebeshka and decoder Zlata demonstrate scores comparable with large baseline LMs on downstream tasks.
That highlights that LMs at a small scale can quickly adapt to the fine-tuning task, though may achieve lower performance in a zero-shot evaluation on BliMP. 
When comparing decoder LMs, we observe that the introduced Zlata outperforms OPT-baseline on paraphrase detection (MRPC \& QQP), entailment/contradiction detection (MNLI), and question answering (BoolQ) downstream tasks.
As for the encoder LMs, the encoder Bebeshka has moderate scores compared to RoBERTa, which, in general, achieves the best scores on GLUE. 
However, Bebeshka outperforms OPT-125M baseline on QQP and MRPC tasks with F1 scores of 73.5\% and 66.4\%, respectively.

The most difficult task for shallow LMs seems to be Recognizing Textual Entailment (RTE). 
We suppose that LMs trained on \textsc{Strict-Small} corpus with an average length of 28.65 tokens (\autoref{tab:corpora-summary}, \autoref{sec:app_results}) or restricted to the 128 maximum sequence length, can perform well on datasets with short sequences and contexts, which can explain lower results on some fine-tuned tasks; another issue can be the fine-tuning hyper-parameters search: perhaps, shallow LMs require more epochs to improve the submitted scores.

\begin{table*}[th!]
\centering
\begin{tabular}{l|ccccc}
\toprule
\textbf{Model} & \textbf{Justice} & \textbf{Deontology} & \textbf{Virtue} & \textbf{Utilitarianism}  & \textbf{Commonsense}\\
\midrule
RoBERTa-large (355M) & \underline{56.7} & \underline{60.3} & 53.0 &  \textbf{79.5}  & \textbf{90.4} \\
GPT-3 few-shot (175B) & 15.2 & 15.9 & 18.2 & \underline{73.7}  & \underline{73.3}  \\
\midrule
Bebeshka (16M)  & \textbf{64.6} & \textbf{71.4} & \textbf{74.1} &  69.0 & - \\
Zlata few-shot (66M)  & 50.7 & 49.6 & \underline{72.0} &  50.3  & 53.3 \\
\bottomrule
\end{tabular}
\caption{\label{tab:ethics_results} 
Accuracy scores on \textsc{ETHICS} benchmark. 
LMs trained on \textsc{Strict-Small} corpus reach results close to the large model baselines reported by \citealp{hendrycks2020aligning}.
We do not report results for the fine-tuning tasks which require the maximum sequence length exceeding the one of an LM. 
The best score is in bold, and the second-best score is underlined.}
\end{table*}
\subsection{Mixed Signals
Generalization}
The MSGS dataset introduced by \cite{warstadt-etal-2020-learning} comprises 20 binary classification tasks and is used to test whether a LM has a preference for linguistic or surface generalizations.  
The evaluation pipeline  of the shared task includes 11 MSGS tasks; we report obtained accuracy scores for the fine-tuned LMs in \autoref{tab:msgs_results}~(\autoref{sec:app_results}). 
The Matthew’s Correlation Coefficient (MCC;~\citealp{matthews1975comparison}) scores suggest that all LMs fine-tuned in a controlled setting show better results (>0.9) than those fine-tuned in an ambiguous scenario, with the only exception for \textit{Control Raising} category; the highest scores are achieved on \textit{Lexical content} and \textit{Relative position} tasks.
\textit{Lexical Content} is a task of classifying sentences with ``the'' (\textit{the mouse} vs \textit{a mouse}) when \textit{Relative Position} is a task of determining whether ``the'' precedes ``a'' in a sentence.
Decoder LMs perform similarly on MSGS tasks chosen for the BabyLM competition, excluding \textit{Syntactic Category}-\textit{Lexical Content} (SC-LC) classification task, where SC is a task of detecting sentences with adjectives. 
A decoder LM Zlata seems to adopt surface generalization during fine-tuning on unambiguous data (SC-LC), whereby the baseline model OPT learns to represent linguistic features.
Bebeshka behaves likewise on the \textit{Syntactic Category} task and reaches scores close to RoBERTa on \textit{Lexical Content} and \textit{Main Verb} classification problems, suggesting that Bebeshka tends to encode surface features.

\subsection{Age of Acquisition}

\citealp{portelance2023predicting} introduced a method for measuring the age-of-acquisition in LMs compared to the actual age-of-acquisition by English American children on words set from the CHILDES corpus.
\autoref{tab:adj_results}~(\autoref{sec:app_results}) illustrates that deviation measured in months for the introduced and baseline LMs. 
The models Zlata and Bebeshka demonstrate comparable scores to the baselines.

\section{Moral Judgments}

\label{sec:moral_judjement}
In this section, we present the results of additional experiments on moral judgements that we conduct outside of the main shared task evaluation. 

We evaluate small-scale LM's understanding of fundamental moral principles in various scenarios covered by \textsc{ETHICS} benchmark~\cite{hendrycks2020aligning}. 
The benchmark consists of 5 morality judgment tasks, including reasonable and fair justice, virtue responses, permitted behaviour depending on context-specified constraints (deontology ethics),  pleasant scenario choice (utilitarianism ethics), and commonsense morality.
We grid search hyper-parameters for our LMs and use test splits for further evaluation.
We fine-tune Bebeshka for ten epochs on each of the tasks and evaluate Zlata in a few-shot setting~(see more details in \autoref{sec:train_details}).
\autoref{tab:ethics_results} outlines the moral judgements classification results.
Our small LMs generally outperform existing baselines with respect to accuracy scores on sentence-level tasks, and the best results are achieved on \textit{Virtue} moral judgements.

We suggest that the efficiency of small LMs in these tasks can be explained by some properties of pre-training data, such as lower mean sequence length, transcribed speech prevalence with single-word reactions or responses, children-directed speech, and imperatives. 
For example, \textit{Virtue} task is a collection of scenario-trait pairs, such as \textit{``Jordan will never do harm to his friends. <sep> caring''}, which have a structure similar to one-word responses in transcribed dialogues.

\section{Conclusion and Future Work}
In this paper, we present our results for the \textsc{Strict-Small} track of the BabyLM competition.
Our submission to the shared task consists of two LMs, namely encoder Bebeshka and decoder Zlata.
We first search for an optimal architecture, minimizing perplexity on the released training corpus, and find that the best models have around 6 encoder layers on average, down from 12 layers of existing base models, and have twice as many attention heads.
When the number of encoder layers fluctuates among the best models, we find that they all have an attention-heads-to-layers ratio of two, which we further use for building our LMs.
Our final LMs, which are scaled-down versions of RoBERTa and GPT-2 with a total of 16M and 66M parameters, perform better than the baseline LMs on development and test BabyLM corpora.  
Zero-shot evaluation results suggest that our shallow LMs have some basic grammatical knowledge of language syntax and morphology.
The introduced LMs also perform better than OPT model on several downstream tasks when having 2 times fewer parameters.
We also observe a good performance of our small LMs in a range of ethics judgment tasks, showing that their vocabulary and after-training knowledge can positively contribute to the morality assessment of the described scenarios. 
These results can serve as baselines for the evaluation of ethical judgment capabilities in small language models.
The achieved scores may be attributed to the interplay between ethical and linguistic rules, particularly in encoding action verbs used to describe moral and immoral behaviour. 
This aspect can be further explored by examining the usage of verbs in various syntactic contexts within the BabyLM corpus and their encoding by trained language models.

In our future work, we plan to determine more capabilities of small LMs, trained on small-size corpora, such as short stories data containing words only 4-year-old children can understand~\cite{eldan2023tinystories}.
We also plan to extend our experiments with an analysis of fine-tuning dynamics to investigate how small models adapt to the tasks.

\section*{Limitations}
Despite achieving good performance on BabyLM test data, our approach has some limitations.
We use a variant of Bayesian optimization (TPE algorithm, \S\ref{sec:model_selection}) to find an optimal range of parameters that we further use for building our LMs.
We predefine constraints for parameters (\autoref{tab:model_selection}) that narrows down the search space and can influence further parameter distributions built with Parzen (kernel density) estimators and, thus, future candidate selection.
Future work can benefit from both expanded search space and parameter limits range. 
The architecture of our small language models, including the number of layers, heads, and hidden layer size, can serve as a minimum lower bound for the parameter search space.

\section*{Acknowledgements}
This work was funded by the ANR project Dik\'{e} (grant
number ANR-21-CE23-0026-02).
\bibliography{anthology,custom}

\begin{thebibliography}{30}
\expandafter\ifx\csname natexlab\endcsname\relax\def\natexlab#1{#1}\fi

\bibitem[{Akiba et~al.(2019)Akiba, Sano, Yanase, Ohta, and
  Koyama}]{10.1145/3292500.3330701}
Takuya Akiba, Shotaro Sano, Toshihiko Yanase, Takeru Ohta, and Masanori Koyama.
  2019.
\newblock \href {https://doi.org/10.1145/3292500.3330701} {Optuna: A
  next-generation hyperparameter optimization framework}.
\newblock In \emph{Proceedings of the 25th ACM SIGKDD International Conference
  on Knowledge Discovery \& Data Mining}, KDD '19, page 2623–2631, New York,
  NY, USA. Association for Computing Machinery.

\bibitem[{Eldan and Li(2023)}]{eldan2023tinystories}
Ronen Eldan and Yuanzhi Li. 2023.
\newblock Tiny{S}tories: How small can language models be and still speak
  coherent english?
\newblock \emph{arXiv preprint arXiv:2305.07759}.

\bibitem[{Fedus et~al.(2022)Fedus, Zoph, and Shazeer}]{fedus2022switch}
William Fedus, Barret Zoph, and Noam Shazeer. 2022.
\newblock Switch transformers: Scaling to trillion parameter models with simple
  and efficient sparsity.
\newblock \emph{The Journal of Machine Learning Research}, 23(1):5232--5270.

\bibitem[{Fu et~al.(2023)Fu, Peng, Ou, Sabharwal, and
  Khot}]{fu2023specializing}
Yao Fu, Hao Peng, Litu Ou, Ashish Sabharwal, and Tushar Khot. 2023.
\newblock Specializing smaller language models towards multi-step reasoning.
\newblock \emph{arXiv preprint arXiv:2301.12726}.

\bibitem[{Hendrycks et~al.(2020)Hendrycks, Burns, Basart, Critch, Li, Song, and
  Steinhardt}]{hendrycks2020aligning}
Dan Hendrycks, Collin Burns, Steven Basart, Andrew Critch, Jerry Li, Dawn Song,
  and Jacob Steinhardt. 2020.
\newblock Aligning {AI} with shared human values.
\newblock \emph{arXiv preprint arXiv:2008.02275}.

\bibitem[{Huebner et~al.(2021)Huebner, Sulem, Cynthia, and
  Roth}]{huebner-etal-2021-babyberta}
Philip~A. Huebner, Elior Sulem, Fisher Cynthia, and Dan Roth. 2021.
\newblock \href {https://doi.org/10.18653/v1/2021.conll-1.49} {{B}aby{BERT}a:
  Learning more grammar with small-scale child-directed language}.
\newblock In \emph{Proceedings of the 25th Conference on Computational Natural
  Language Learning}, pages 624--646, Online. Association for Computational
  Linguistics.

\bibitem[{Jia et~al.(2019)Jia, Tillman, Maggioni, and
  Scarpazza}]{jia2019dissecting}
Zhe Jia, Blake Tillman, Marco Maggioni, and Daniele~Paolo Scarpazza. 2019.
\newblock Dissecting the graphcore ipu architecture via microbenchmarking.
\newblock \emph{arXiv preprint arXiv:1912.03413}.

\bibitem[{Kaplan et~al.(2020)Kaplan, McCandlish, Henighan, Brown, Chess, Child,
  Gray, Radford, Wu, and Amodei}]{kaplan2020scaling}
Jared Kaplan, Sam McCandlish, Tom Henighan, Tom~B Brown, Benjamin Chess, Rewon
  Child, Scott Gray, Alec Radford, Jeffrey Wu, and Dario Amodei. 2020.
\newblock Scaling laws for neural language models.
\newblock \emph{arXiv preprint arXiv:2001.08361}.

\bibitem[{Liu et~al.(2019)Liu, Ott, Goyal, Du, Joshi, Chen, Levy, Lewis,
  Zettlemoyer, and Stoyanov}]{liu2019roberta}
Yinhan Liu, Myle Ott, Naman Goyal, Jingfei Du, Mandar Joshi, Danqi Chen, Omer
  Levy, Mike Lewis, Luke Zettlemoyer, and Veselin Stoyanov. 2019.
\newblock {RoBERTa: A Robustly Optimized BERT Pretraining Approach}.
\newblock \emph{arXiv preprint arXiv:1907.11692}.

\bibitem[{Mahowald et~al.(2023)Mahowald, Ivanova, Blank, Kanwisher, Tenenbaum,
  and Fedorenko}]{mahowald2023dissociating}
Kyle Mahowald, Anna~A Ivanova, Idan~A Blank, Nancy Kanwisher, Joshua~B
  Tenenbaum, and Evelina Fedorenko. 2023.
\newblock Dissociating language and thought in large language models: a
  cognitive perspective.
\newblock \emph{arXiv preprint arXiv:2301.06627}.

\bibitem[{Matthews(1975)}]{matthews1975comparison}
Brian~W Matthews. 1975.
\newblock Comparison of the predicted and observed secondary structure of t4
  phage lysozyme.
\newblock \emph{Biochimica et Biophysica Acta (BBA)-Protein Structure},
  405(2):442--451.

\bibitem[{Moi and Patry(2023)}]{Moi_HuggingFace_s_Tokenizers_2023}
Anthony Moi and Nicolas Patry. 2023.
\newblock \href {https://github.com/huggingface/tokenizers} {{HuggingFace's
  Tokenizers}}.

\bibitem[{Portelance et~al.(2023)Portelance, Duan, Frank, and
  Lupyan}]{portelance2023predicting}
Eva Portelance, Yuguang Duan, Michael~C. Frank, and Gary Lupyan. 2023.
\newblock Predicting age of acquisition for children’s early vocabulary in
  five languages using language model surprisal.

\bibitem[{Puvis~de Chavannes et~al.(2021)Puvis~de Chavannes, Kongsbak, Rantzau,
  and Derczynski}]{puvis-de-chavannes-etal-2021-hyperparameter}
Lucas~H{\o}yberg Puvis~de Chavannes, Mads Guldborg~Kjeldgaard Kongsbak, Timmie
  Rantzau, and Leon Derczynski. 2021.
\newblock \href {https://doi.org/10.18653/v1/2021.sustainlp-1.12}
  {Hyperparameter power impact in transformer language model training}.
\newblock In \emph{Proceedings of the Second Workshop on Simple and Efficient
  Natural Language Processing}, pages 96--118, Virtual. Association for
  Computational Linguistics.

\bibitem[{Radford et~al.(2019)Radford, Wu, Child, Luan, Amodei, Sutskever
  et~al.}]{radford2019language}
Alec Radford, Jeffrey Wu, Rewon Child, David Luan, Dario Amodei, Ilya
  Sutskever, et~al. 2019.
\newblock Language models are unsupervised multitask learners.
\newblock \emph{OpenAI blog}, 1(8):9.

\bibitem[{Sanh et~al.(2019)Sanh, Debut, Chaumond, and
  Wolf}]{sanh2019distilbert}
Victor Sanh, Lysandre Debut, Julien Chaumond, and Thomas Wolf. 2019.
\newblock Distil{BERT}, a distilled version of {BERT}: smaller, faster, cheaper
  and lighter.
\newblock \emph{arXiv preprint arXiv:1910.01108}.

\bibitem[{Schick and Sch{\"u}tze(2021)}]{schick-schutze-2021-just}
Timo Schick and Hinrich Sch{\"u}tze. 2021.
\newblock \href {https://doi.org/10.18653/v1/2021.naacl-main.185} {It{'}s not
  just size that matters: Small language models are also few-shot learners}.
\newblock In \emph{Proceedings of the 2021 Conference of the North American
  Chapter of the Association for Computational Linguistics: Human Language
  Technologies}, pages 2339--2352, Online. Association for Computational
  Linguistics.

\bibitem[{Schramowski et~al.(2022)Schramowski, Turan, Andersen, Rothkopf, and
  Kersting}]{schramowski2022large}
Patrick Schramowski, Cigdem Turan, Nico Andersen, Constantin~A Rothkopf, and
  Kristian Kersting. 2022.
\newblock Large pre-trained language models contain human-like biases of what
  is right and wrong to do.
\newblock \emph{Nature Machine Intelligence}, 4(3):258--268.

\bibitem[{Sennrich et~al.(2016)Sennrich, Haddow, and
  Birch}]{sennrich-etal-2016-neural}
Rico Sennrich, Barry Haddow, and Alexandra Birch. 2016.
\newblock \href {https://doi.org/10.18653/v1/P16-1162} {Neural machine
  translation of rare words with subword units}.
\newblock In \emph{Proceedings of the 54th Annual Meeting of the Association
  for Computational Linguistics (Volume 1: Long Papers)}, pages 1715--1725,
  Berlin, Germany. Association for Computational Linguistics.

\bibitem[{Shoeybi et~al.(2019)Shoeybi, Patwary, Puri, LeGresley, Casper, and
  Catanzaro}]{shoeybi2019megatron}
Mohammad Shoeybi, Mostofa Patwary, Raul Puri, Patrick LeGresley, Jared Casper,
  and Bryan Catanzaro. 2019.
\newblock Megatron-lm: {T}raining multi-billion parameter language models using
  model parallelism.
\newblock \emph{arXiv preprint arXiv:1909.08053}.

\bibitem[{Tao et~al.(2022)Tao, Hou, Zhang, Shang, Jiang, Liu, Luo, and
  Wong}]{tao-etal-2022-compression}
Chaofan Tao, Lu~Hou, Wei Zhang, Lifeng Shang, Xin Jiang, Qun Liu, Ping Luo, and
  Ngai Wong. 2022.
\newblock \href {https://doi.org/10.18653/v1/2022.acl-long.331} {Compression of
  generative pre-trained language models via quantization}.
\newblock In \emph{Proceedings of the 60th Annual Meeting of the Association
  for Computational Linguistics (Volume 1: Long Papers)}, pages 4821--4836,
  Dublin, Ireland. Association for Computational Linguistics.

\bibitem[{Tucker et~al.(2022)Tucker, Eisape, Qian, Levy, and
  Shah}]{tucker-etal-2022-syntax}
Mycal Tucker, Tiwalayo Eisape, Peng Qian, Roger Levy, and Julie Shah. 2022.
\newblock \href {https://doi.org/10.18653/v1/2022.naacl-main.394} {When does
  syntax mediate neural language model performance? evidence from dropout
  probes}.
\newblock In \emph{Proceedings of the 2022 Conference of the North American
  Chapter of the Association for Computational Linguistics: Human Language
  Technologies}, pages 5393--5408, Seattle, United States. Association for
  Computational Linguistics.

\bibitem[{van Schijndel et~al.(2019)van Schijndel, Mueller, and
  Linzen}]{van-schijndel-etal-2019-quantity}
Marten van Schijndel, Aaron Mueller, and Tal Linzen. 2019.
\newblock \href {https://doi.org/10.18653/v1/D19-1592} {Quantity doesn{'}t buy
  quality syntax with neural language models}.
\newblock In \emph{Proceedings of the 2019 Conference on Empirical Methods in
  Natural Language Processing and the 9th International Joint Conference on
  Natural Language Processing (EMNLP-IJCNLP)}, pages 5831--5837, Hong Kong,
  China. Association for Computational Linguistics.

\bibitem[{Wang et~al.(2018)Wang, Singh, Michael, Hill, Levy, and
  Bowman}]{wang-etal-2018-glue}
Alex Wang, Amanpreet Singh, Julian Michael, Felix Hill, Omer Levy, and Samuel
  Bowman. 2018.
\newblock \href {https://doi.org/10.18653/v1/W18-5446} {{GLUE}: A multi-task
  benchmark and analysis platform for natural language understanding}.
\newblock In \emph{Proceedings of the 2018 {EMNLP} Workshop {B}lackbox{NLP}:
  Analyzing and Interpreting Neural Networks for {NLP}}, pages 353--355,
  Brussels, Belgium. Association for Computational Linguistics.

\bibitem[{Warstadt and Bowman(2022)}]{warstadt2022artificial}
Alex Warstadt and Samuel~R Bowman. 2022.
\newblock What artificial neural networks can tell us about human language
  acquisition.
\newblock \emph{Algebraic Structures in Natural Language}, pages 17--60.

\bibitem[{Warstadt et~al.(2023)Warstadt, Mueller, Choshen, Wilcox, Zhuang,
  Ciro, Mosquera, Williams, Paranjabe, Linzen, and
  Cotterell}]{warstadt-et-al-2023-babylm}
Alex Warstadt, Aaron Mueller, Leshem Choshen, Ethan~Gotlieb Wilcox, Chengxu
  Zhuang, Juan Ciro, Rafael Mosquera, Adina Williams, Bhargavi Paranjabe, Tal
  Linzen, and Ryan Cotterell. 2023.
\newblock Findings of the 2023 {B}aby{LM} {C}hallenge: {S}ample-efficient
  pretraining on developmentally plausible corpora.
\newblock In \emph{Proceedings of the 2023 {B}aby{LM} {C}hallenge}. Association
  for Computational Linguistics (ACL).

\bibitem[{Warstadt et~al.(2020{\natexlab{a}})Warstadt, Parrish, Liu, Mohananey,
  Peng, Wang, and Bowman}]{warstadt-etal-2020-blimp-benchmark}
Alex Warstadt, Alicia Parrish, Haokun Liu, Anhad Mohananey, Wei Peng, Sheng-Fu
  Wang, and Samuel~R. Bowman. 2020{\natexlab{a}}.
\newblock \href {https://doi.org/10.1162/tacl_a_00321} {{BL}i{MP}: The
  benchmark of linguistic minimal pairs for {E}nglish}.
\newblock \emph{Transactions of the Association for Computational Linguistics},
  8:377--392.

\bibitem[{Warstadt et~al.(2020{\natexlab{b}})Warstadt, Zhang, Li, Liu, and
  Bowman}]{warstadt-etal-2020-learning}
Alex Warstadt, Yian Zhang, Xiaocheng Li, Haokun Liu, and Samuel~R. Bowman.
  2020{\natexlab{b}}.
\newblock \href {https://doi.org/10.18653/v1/2020.emnlp-main.16} {Learning
  which features matter: {R}o{BERT}a acquires a preference for linguistic
  generalizations (eventually)}.
\newblock In \emph{Proceedings of the 2020 Conference on Empirical Methods in
  Natural Language Processing (EMNLP)}, pages 217--235, Online. Association for
  Computational Linguistics.

\bibitem[{Wu et~al.(2022)Wu, Geiger, Rozner, Kreiss, Lu, Icard, Potts, and
  Goodman}]{wu-etal-2022-causal}
Zhengxuan Wu, Atticus Geiger, Joshua Rozner, Elisa Kreiss, Hanson Lu, Thomas
  Icard, Christopher Potts, and Noah Goodman. 2022.
\newblock \href {https://doi.org/10.18653/v1/2022.naacl-main.318} {Causal
  distillation for language models}.
\newblock In \emph{Proceedings of the 2022 Conference of the North American
  Chapter of the Association for Computational Linguistics: Human Language
  Technologies}, pages 4288--4295, Seattle, United States. Association for
  Computational Linguistics.

\bibitem[{Yao et~al.(2021)Yao, Huang, Wang, Dong, and
  Wei}]{yao-etal-2021-adapt}
Yunzhi Yao, Shaohan Huang, Wenhui Wang, Li~Dong, and Furu Wei. 2021.
\newblock \href {https://doi.org/10.18653/v1/2021.findings-acl.40}
  {Adapt-and-distill: Developing small, fast and effective pretrained language
  models for domains}.
\newblock In \emph{Findings of the Association for Computational Linguistics:
  ACL-IJCNLP 2021}, pages 460--470, Online. Association for Computational
  Linguistics.

\end{thebibliography}
\bibliographystyle{acl_natbib}

\appendix
\newpage 
\clearpage

\onecolumn 
\section{Experimental Framework}

\label{sec:experimental_frame}

\begin{table}[th!]
\small
\centering
\begin{tabular}{lrrrr} 
\toprule
\textbf{Dataset} & \textbf{\# Sentences} &  \textbf{Avg. length$^*$}
& \textbf{Questions (Proportion)} & \textbf{Proportion}  \\
\midrule
CHILDES & 64258 & 7.17 & 39\% &  5\%\\
British National Corpus (BNC) & 66100 & 16.06 & 17\% & 8\%\\
Children’s Book Test & 25946  & 25.49 & 3\% & 6\% \\
Children’s Stories Text Corpus & 5569 & 60.58 & 1\% & 3\% \\
Standardized Project Gutenberg Corpus & 90402 & 16.22 & 0\% & 10\% \\
OpenSubtitles & 417984 & 9.94 & 17\% & 31\%  \\
QCRI Educational Domain Corpus & 91904 & 16.38 & 0\% & 11\% \\
Wikipedia & 40876 & 51.28 & 0\% & 10\% \\
Simple Wikipedia & 9938 & 14.57 & 6\% & 15\% \\
Switchboard Dialog Act Corpus & 5569 & 60.58 & 0\% & 1\%\\
\midrule
Total & 832274 & 28.65 & 13.1\% & 100\%\\
\bottomrule
\end{tabular}
\caption{Statistics of the training corpus offered in the \textsc{Strict-Small} track of BabyLM competition. $^*$= Average tokenized text length.  
}
\label{tab:corpora-summary}
\end{table}

\section{Tokenization Tests}
\label{sec:tokenization_tests}
We compare the tokenization of Bebeshka and RoBERTa on the corpus of \textsc{Strict-Small} track and find that the tokenization coincides on 87\% of the sequences.   
We manually analyse a random sample of 100 non-matching tokenization cases and find that those fall on transcribed speech sentences with no more than three words or include two words missing in RoBERTa vocabulary but processed as a whole word by Bebeshka LM (\textit{sweetie} and \textit{duke}). 
We also found that the RoBERTa tokenizer splits non-capitalised first names or other terms used for addressing (\textit{th-omas}, \textit{m-ister}, \textit{mom-my}) opposed to Bebeshka. 
\section{Training Details}
\label{sec:train_details}

\subsection{Pre-training parameters}
We experimented with the same configuration for our decoder LM Zlata as we used for Bebeshka, including 4 layers and the same type of positional embeddings; however, that always resulted in gradients underflow and that loss was not decreasing. 
We manually found the 6-layer and absolute positional embedding configurations by increasing and traversing values of the parameters that were grid searched for Bebeshka (\autoref{tab:model_selection}). 
We pre-train our LMs using 4x IPUs freely available in Paperspace\footnote{\url{https://www.paperspace.com}} and use IPU Trainer API. 
We use auto-loss scaling with an initial value of 16384 and half-precision for training our LMs.
Training with IPUs requires specifying IPU configuration, containing instructions for mapping layers between the devices; for Bebeshka, we use one layer per IPU, and for Zlata, we use that parameter equal to 2.
For both LMs, we use per-device training batch size equal to 1 and gradient accumulation steps equal to 64.
Each batch consists of 1,000 concatenated data examples from the training corpus. 
The time for the computational graph construction took under 10 minutes for both training both LMs.
                                    
\subsection{Fine-tuning parameters}
\paragraph{BabyLM Evaluation}
For Bebeshka fine-tuning, we use parameters used by default in the evaluation pipeline of the competition, that is, learning rate equal to 5e-5, batch size equal to 64, and maximum epochs equal to 10. 
For Zlata fine-tuning, we use the learning rate equal to 1e-4 and fine-tune the tasks for 5 epochs. That allowed us to reduce fine-tuning time.
Note that the performance of our LMs can be improved upon the submitted results if grid search the optimal hyper-parameters.
\paragraph{Moral Judgement}
We use a weighted loss for fine-tuning Bebeshka and grid search optimal parameters using an official implementation by the authors of the dataset.\footnote{\url{https://github.com/hendrycks/ethics}}
For our GPT-2 based model Zlata, we use an existing evaluation harness benchmark in the k-shot setting with k equal to 15.\footnote{\url{https://github.com/EleutherAI/lm-evaluation-harness/}} 

\section{Evaluation Results}
\label{sec:app_results}
\begin{table*}[th!]
\centering
\small
\resizebox{0.96\textwidth}{!}{
\begin{tabular}{lllllllllllll} 
\toprule
\textbf{Model} & \rotatebox{60}{\textbf{Anaphor Agr.}}  & \rotatebox{60}{\textbf{Arg. Structure}}  & \rotatebox{60}{\textbf{Binding}}  & \rotatebox{60}{\textbf{Control/Raising}}  & \rotatebox{60}{\textbf{D-N Agr.}}  & \rotatebox{60}{\textbf{Ellipsis}}  & \rotatebox{60}{\textbf{Filler-Gap}}  & \rotatebox{60}{\textbf{Irregular Forms}}  & \rotatebox{60}{\textbf{Island Effects}}  & \rotatebox{60}{\textbf{NPI Licensing}}  & \rotatebox{60}{\textbf{Quantifiers}}  & \rotatebox{60}{\textbf{S-V Agr.}}  \\
\midrule 
OPT-125M  & 63.8  & \textbf{70.6}  & 67.1  & \underline{66.5}  & 78.5  & 62.0 & 63.8  & 67.5  & \textbf{48.6}  & 46.7  & 59.6  & 56.9  \\
RoBERTa-base  & \textbf{81.5}  & 67.1  & \underline{67.3}  & \textbf{67.9}  & \textbf{90.8}  & \textbf{76.4}  & 63.5  & 87.4 & 39.9  & \textbf{55.9}  & \textbf{70.5}  & \textbf{65.4}  \\
T5-base  & 68.9  & 63.8  & 60.4  & 60.9  & 72.2  & 34.4  & 48.2  & 77.6  & \underline{45.6}  & 47.8  & 61.2  & \underline{65.0} \\
\midrule 
Bebeshka  & 52.0  & 57.3  & 61.5 & 56.8  & 58.0 & 37.9  & \underline{64.7}  & 84.5  & 44.8 & 39.2  & 49.7  & 53.2 \\
Zlata & 72.0  & \underline{68.1} & 66.9 & 61.7  & 80.0  & 48.6  & \textbf{65.4}  & \underline{92.1} & 40.3 & \underline{50.4}  & 66.4  & 60.3 \\
\midrule 
Bebeshka-2  & \underline{77.7}  & 60.2  & \textbf{68.0} & 56.2  & \underline{87.4} & \underline{68.8}  & \underline{64.7}  & \textbf{92.8}  & 37.0 & 45.1  & \underline{70.2}  & 60.5 \\
\bottomrule
\end{tabular}}
\caption{
Model evaluation results on BLiMP dataset.
The scores show the model's accuracy in distinguishing between the grammatical and ungrammatical sentences within each minimal pair.
The best score is in bold, and the second-best score is underlined.}
\label{tab:blimp_results}
\end{table*}

\begin{table*}[th!]
\centering
\footnotesize
\begin{tabular}{l|lllll|llllll}
\toprule
\multirow{2}{*}{\textbf{Model}} & \textbf{CR}  & \textbf{LC}  & \textbf{MV}  & \textbf{RP}  & \textbf{SC}  & \textbf{CR LC}  & \textbf{CR RTP}  & \textbf{MV LC}  & \textbf{MV RTP}  & \textbf{SC LC}  & \textbf{SC RP}  \\
\cmidrule{2-12}
& \multicolumn{5}{c|}{\textbf{Control}} & \multicolumn{6}{c}{\textbf{Ambiguous}}  \\
\midrule
OPT & \textbf{50.8}  & 53.6  & \textbf{99.5}  & \textbf{99.9}  & 77.2  & \textbf{0.4}  & -70.3  & \underline{-72.1}  & -77.6  & \underline{13.8}  & -68.9  \\
RoBERTa & 43.1  & \textbf{100.0} & 97.7  & 76.7  & \textbf{86.2}  & -28.3  & -77.7  & -99.3  & -79.4  & \textbf{16.3}  & -45.0  \\
T5  & 21.1 & \textbf{100.0}   & 33.4  & 82.5  & \underline{77.6}  & -78.3  & \textbf{-62.0}  & -100.0  & -79.7  & -25.3 & \underline{-39.4}  \\
\midrule
Bebeshka & 13.0 & \textbf{100.0}   & 97.0  & 72.0 & 41.0  & -95.0 & \underline{-63.0}  & -100.0 & \underline{-66.0}  & -58.0  & -62.0 \\
Zlata & 37.0 & \underline{79.0}  & 90.0 & 87.0  & 64.0 & \underline{-9.0}  & -85.0  & \textbf{-70.0}   & -94.0  & -58.0  & \textbf{-39.0}  \\
\midrule
Bebeshka-2 & \underline{49.4} & \textbf{100.0}   & \underline{98.2}  & \underline{88.3} & 61.5  & -28.9 & -80.4  & -100.0 & \textbf{-40.8}  & -57.2  & -46.4 \\
\bottomrule
\end{tabular}
\caption{
Model evaluation results: Matthews Correlation Coefficient (MCC) on the synthetic MSGS dataset, multiplied by 100.
CR=\textit{Control Raising}, LC=\textit{Lexical Content}, MV=\textit{Main Verb}, RP=\textit{Relative Position}, SC=\textit{Syntactic Category},  RTP=\textit{Relative Token Position}. 
Control columns correspond to the control experiments when an LM is trained to classify sentences with certain linguistic and surface features.
Ambiguous correspond to the experiments when an LM is tested on a single-feature dataset (for example, LC) after training on a set with labels consistent across both linguistic and surface features (SC LC). 
The highest score is in bold, and the second-highest score is underlined.}
\label{tab:msgs_results}
\end{table*}

\begin{table*}[th!]
\centering
\small
\begin{tabular}{lcccc}\\ 
\toprule
\textbf{Model} & \textbf{Overall (591 words)}  & \textbf{Nouns (322)}  & \textbf{Predicates (167)}  & \textbf{Function words (102)}  \\
\midrule
OPT-125M  & 2.03  & 1.98  & 1.81  & 2.57 \\
RoBERTa-base  & 2.06  & 1.99  & 1.85  & 2.65\\
T5-base  & 2.04  & 1.97  & 1.82  & 2.64  \\
\midrule
Bebeshka  & 2.06  & 1.98  & 1.84  & 2.66\\
Zlata  & 2.07  & 1.99  & 1.83  & 2.67  \\

\bottomrule
\end{tabular}\\
\caption{
Age-of-acquisition (AoA) predictions on child-directed utterances from CHILDES data. 
The scores are Mean Absolute Deviation scores in months between the actual average AoA of the words by  American English-speaking children and model predicted AoA, measured as a likelihood of the words' usage across all the contexts (surprisal scores). The lower the MAD scores, the better.
Top-5 words with the highest surprisal scores for LMs: Zlata: 
\textit{snowsuit}, \textit{applesauce}, \textit{lawn mower}, \textit{sprinkler}, \textit{tricycle}; 
Bebeshka: 
\textit{snowsuit}, \textit{hen}, \textit{turkey}, \textit{belt}, \textit{lamb}.
}
\label{tab:adj_results}
\end{table*}

\end{document}